\documentclass{article}
\usepackage[final]{neurips_2024}
\usepackage[utf8]{inputenc} 
\usepackage[T1]{fontenc}   
\usepackage{hyperref}       
\usepackage{url}            
\usepackage{booktabs}       
\usepackage{amsfonts}       
\usepackage{nicefrac}       
\usepackage{microtype}      
\usepackage{xcolor}         
\usepackage{multirow}
\usepackage{wrapfig} 
\usepackage{graphicx}
\usepackage{subcaption} 
\usepackage{bm}
\usepackage{amsopn}
\usepackage{amsmath}

\title{3D Focusing-and-Matching Network for Multi-Instance Point Cloud Registration}

\author{
  Liyuan Zhang, Le Hui$^*$, Qi Liu, Bo Li, Yuchao Dai$^*$\\
  School of Electronics and Information, Northwestern Polytechnical University\\
  Shaanxi Key Laboratory of Information Acquisition and Processing\\
  \small\texttt{zhangliyuannpu@mail.nwpu.edu.cn, \{huile, liuqi, libo, daiyuchao\}@nwpu.edu.cn} \\
}

\begin{document}

\maketitle
\let\thefootnote\relax\footnotetext{$^*$Corresponding authors.}
\let\thefootnote\relax\footnotetext{Liyuan Zhang, Le Hui, Qi Liu, Bo Li, and Yuchao Dai are with the Key Lab of Shaanxi Key Laboratory of
Information Acquisition and Processing, School of Electronics And Information, Northwestern Polytechnical University, China.}

\begin{abstract}
Multi-instance point cloud registration aims to estimate the pose of all instances of a model point cloud in the whole scene. Existing methods all adopt the strategy of first obtaining the global correspondence and then clustering to obtain the pose of each instance. However, due to the cluttered and occluded objects in the scene, it is difficult to obtain an accurate correspondence between the model point cloud and all instances in the scene. To this end, we propose a simple yet powerful 3D focusing-and-matching network for multi-instance point cloud registration by learning the multiple pair-wise point cloud registration. Specifically, we first present a 3D multi-object focusing module to locate the center of each object and generate object proposals. By using self-attention and cross-attention to associate the model point cloud with structurally similar objects, we can locate potential matching instances by regressing object centers. Then, we propose a 3D dual-masking instance matching module to estimate the pose between the model point cloud and each object proposal. It performs instance mask and overlap mask masks to accurately predict the pair-wise correspondence. Extensive experiments on two public benchmarks, Scan2CAD and ROBI, show that our method achieves a new state-of-the-art performance on the multi-instance point cloud registration task. The project page is at \url{https://npucvr.github.io/3DFMNet/}.
\end{abstract}

\section{Introduction}
Point cloud registration, a fundamental process in computer vision, involves aligning two point clouds through estimating a rigid transformation. In practical applications like robotic bin picking, multi-instance registration emerges as a critical need, demanding the alignment of a model's point cloud with multiple instances within the scene. This task presents heightened complexity compared to single-point cloud registration, primarily due to challenges such as the uncertain number of instances and inter-instance occlusions. These complexities are particularly pronounced in cluttered environments, where precise alignment becomes pivotal for effective robotic operations. Therefore, how to improve the accuracy of multi-instance point cloud registration is still a challenging issue. 

\begin{figure}[t]
\centering
\includegraphics[width=1.0\textwidth]{"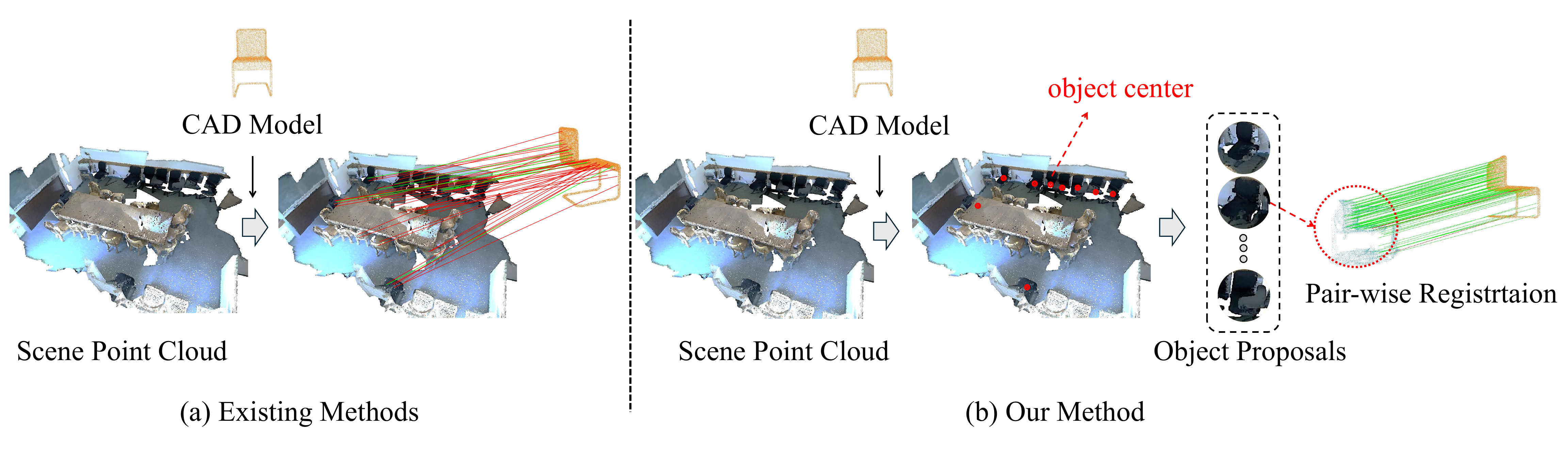"}
\caption{Comparison between our method and existing methods in multi-instance point cloud registration. Our method decomposes the multi-instance point cloud registration into multiple pair-wise point cloud registration.}
\label{fig:motivation}
\end{figure}

There are a few efforts for tackling multi-instance point cloud registration. Existing pipelines can be roughly divided into two types: two-stage and one-stage. For the two-stage process, we first extract point correspondences between the model point cloud and scene point clouds, and then recover per-instance transformations through multi-model fitting~\cite{kluger2020consac,tang2022multi,yuan2022pointclm}. Although two-stage methods are simple and feasible, the success of these methods largely depends on the quality of the correspondence. Furthermore, due to cluttered and occluded objects, it is still difficult to accurately cluster the correspondences into individual instances for subsequent pair-wise registration. For the one-stage process, it takes the model point cloud and scene point cloud as inputs, and directly outputs pose. As a representative one-stage work, Yu~\emph{et al.}\cite{yu2024learning} proposed a coarse-to-fine framework, which learns to extract instance-aware correspondences for estimating transformations without multi-model fitting. Due to the consideration of instance-level information in correspondence, it can obtain fine-grained features, thereby boosting the performance. However, for the scene with multiple objects, obtaining accurate instance-level correspondence is very difficult, especially for the cluttered and occluded objects. Therefore, to alleviate the difficulty of learning correspondence between the model point cloud and multiple objects in the scene, as shown in Figure~\ref{fig:motivation}, we consider first focusing on the object centers, and then learning the matching between the object proposal and the model point cloud.

In this paper, we propose a simple yet powerful 3D focusing-and-matching network for multi-instance point cloud registration. The core idea of our method is to decompose the multi-instance point cloud registration into multiple pair-wise point cloud registrations. Specifically, we propose a 3D multi-object focusing module to localize the potential object centers and generate object proposals. To associate the object with the input CAD model, we use self-attention and cross-attention to learn the structurally similar features, thereby improving the accuracy of prediction for object centers. Based on the learned object center, we incorporate the radius of the CAD model to generate object proposals through ball query operation. After that, we propose a 3D dual-masking instance matching module to learn accurate pair-wise registration between the CAD model and object proposal. It adopts an instance mask to filter the background points in the object proposal and uses an overlap mask to improve the pair-wise partial registration of incomplete objects.

In summary, our contributions lie in three aspects:

\begin{enumerate}
    \item Our primary contribution does not lie in the network architecture but rather in proposing a new pipeline to address the multi-instance point cloud registration problem. Existing methods (such as PointCLM~\cite{yuan2022pointclm} and MIRETR~\cite{yu2024learning}) mainly learn correspondence between the one CAD model and multiple objects (one-to-many paradigm), while our method decompose the one-to-many paradigm into multiple pair-wise point cloud registration (multiple one-to-one paradigm) by first detecting the object centers and then learning the matching between the CAD model and each object proposal.
    \vspace{-3pt}
    \item Our new pipeline is simple yet powerful, achieving the new state-of-the-art on both Scan2CAD~\cite{avetisyan2019scan2cad} and ROBI~\cite{yang2021robi} datasets. Especially on the challenging ROBI dataset, our method significantly outperforms the previous SOTA MIRETR by about 7\% in terms of MR, MP, and MF.
    \vspace{-3pt}
    \item The progressive decomposition approach of transforming multi-instance point cloud registration into multiple pair-wise registrations, as proposed in our paper, also holds significant insights for other tasks, such as multi-target tracking and map construction.
\end{enumerate}

\section{Related Work}
\textbf{Point Cloud Registration} is a crucial task in fields such as robotics and autonomous driving, which usually involving three stages: point matching, outlier rejection, and pose estimation. Acquiring accurate point correspondences is critical for successful registration, making the first two stages particularly important. Accurate point matching deeply relies on features that are descriptive and rotation invariant. Many researchers have made efforts on this, including handcrafted descriptors\cite{FPFH,ERCVPR2010} and learning-based descriptors\cite{3dmatch,fcgf,perfectmatch,ppfnet,predator,d3feat,buffer}. Some recent coarse-to-fine frameworks\cite{CofiNet,qin2022geometric} bypass keypoint detection and achieve accurate correspondences in large-scale scenes. To handle the problem of outliers, RANSAC\cite{RANSAC} and its variants\cite{RANSACviccv,graphransac} ) follow the hypothesis-and-verification process to reject outliers. And some learning-based methods\cite{dgr,3dregnet,pointdsc, shen2022reliable} for eliminating outliers have also been proposed for robust pose estimation. On the other hand, several methods\cite{dcp,omnet,fmr,pointnetlk,prnet,rpcgf,rpmnet} directly estimate the transformation with a neural network in an end-to-end manner. However, these point cloud registration methods almost focus on the one-to-one problem which only need to solve the transformation between two point clouds. So they cannot directly work when faced with the challenge of a large number of instances in multi-instance registration tasks and heavy intra-instance occlusion.

\textbf{Multi-Instance Point Cloud Registration}, which aligns a source point cloud to its various instances within a target point cloud, has received relatively less attention. Unlike multi-way registration\cite{multi-wayreg}, which aims to create a globally consistent reconstruction from multiple fragments through pairwise registration\cite{FGR}, multi-instance registration involves not only rejecting outliers from noisy correspondences but also identifying the inlier set for each individual instance. This makes it even more challenging than the traditional registration problem. Early methods of multi-model fitting were used for this task. In the early stages of this task's development, various methods of multi-model fitting were employed. RANSAC-based approaches\cite{kluger2020consac,Progressive-X,Progressive-X+,BMVC2004} followed a hypothesis verification approach to fit multiple models, while another method, based on clustering\cite{magri2014t,magri2016multiple,lowrank,tang2022multi,yuan2022pointclm,cao2024instance,yu2024efficient}, entailed sampling an extensive set of hypotheses and clustering the correspondences based on their residuals under these hypotheses. Recently, \cite{yu2024learning} proposed an instance-aware correspondence extraction method for end-to-end multi-instance pose estimation. However, existing methods are often affected by outliers from other instances and highly rely on scene-specific global feature extraction. In this approach, the multi-instance registration problem is addressed by detecting matching instance centers in the scene and subsequently splitting the instances, effectively transforming it into a pairwise point cloud registration problem.

\begin{figure}[h]
    \centering
    \includegraphics[width=\textwidth]{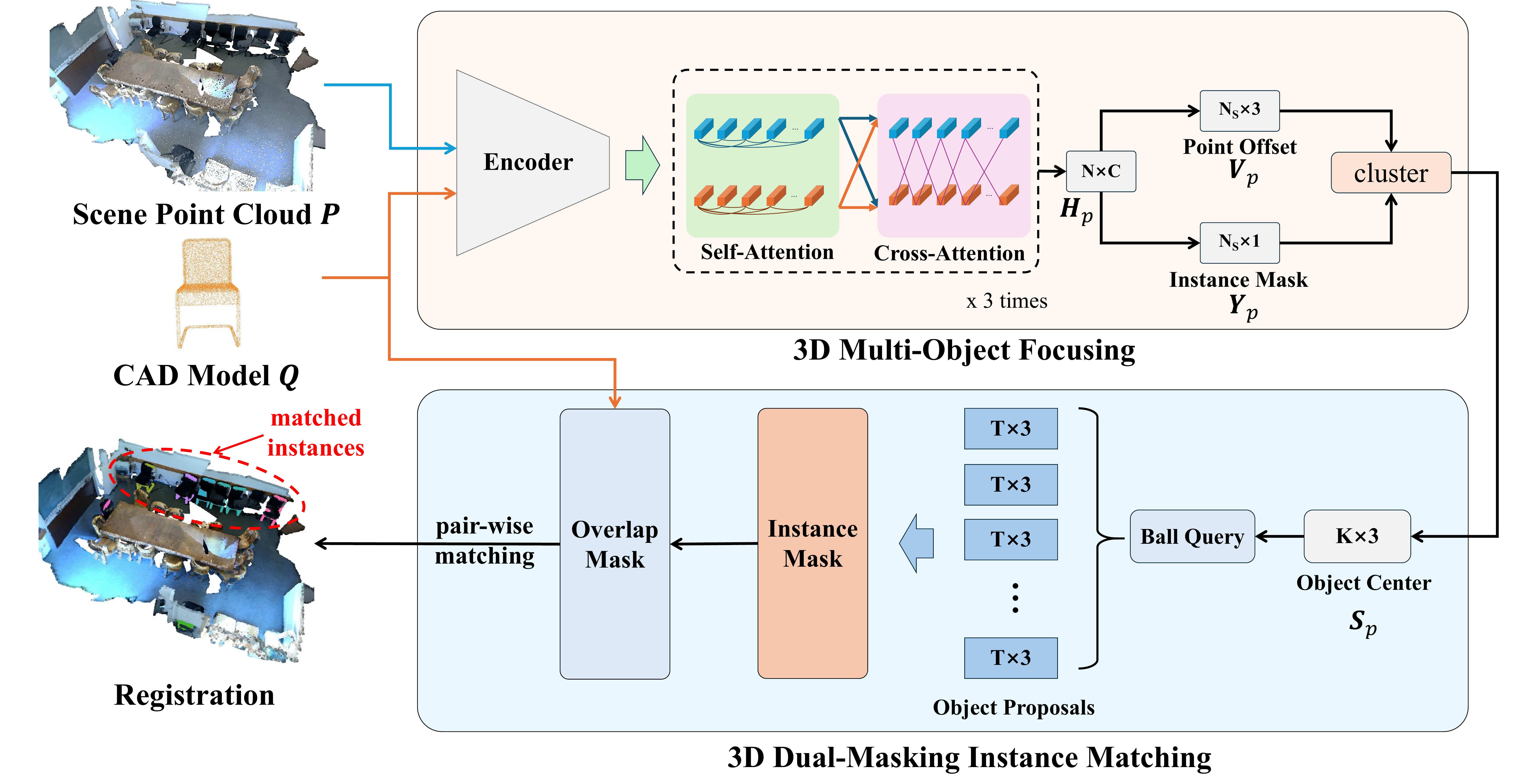}
    \vspace{-15pt}
    \caption{The framework of our 3D focusing-and-matching network for multi-instance pint cloud registration. Given the scene point cloud and the CAD model, we first present the 3D multi-object focusing module to localize the centers of the potential objects in the scene. Then, we design the 3D dual-masking instance matching module to learn pair-wise point cloud registration from the localized object proposals.}
    \label{fig:framework}
\end{figure}

\section{Method}
The overall pipeline is illustrated in Figure~\ref{fig:framework}. Our method is a two-stage framework, which first localizes the center of each object and then performs pair-wise correspondence. For the first stage, we present a 3D multi-object focusing module (Sec.~\ref{sec:sec_3dfocus}) for detecting the potential instance centers by learning the correlation between the input model point cloud and the scene point cloud. For the second stage, we design a 3D dual-masking instance matching module (Sec.~\ref{sec:sec_3dmatch}) for predicting pair-wise correspondence between the input model point cloud and the localized region of each object center. At last, we introduce the loss functions of our method (Sec.~\ref{sec:sec_loss}).

\subsection{3D Multi-Object Focusing}\label{sec:sec_3dfocus}
As the first stage of our method, the 3D multi-object focusing module aims to regress the center of the potential objects for generating high-quality proposals for pair-wise correspondence. Compared to predicting the bounding box or mask of an instance, directly regressing the center of the object is much easier, especially in cluttered and occluded scenes. In order to accurately detect the object center, we first learn the correlation between the model point cloud and the scene point cloud. Then, we predict the object center by learning the offset of each point. Finally, we introduce how to construct 3D object proposals for subsequent pair-wise point cloud registration.

\textbf{Feature correlation learning.} We design a simple yet efficient feature extraction structure to learn the correlation between the scene point cloud $\bm{P}\in\mathbb{R}^{N\times 3}$ and the model point cloud $\bm{Q}\in\mathbb{R}^{M\times 3}$, where $N$ and $M$ are the numbers of their points, respectively. Note that for a fair comparison, we do not use RGB information of point cloud. Before learning correlation, we adopt the encoder of KPConv~\cite{thomas2019kpconv} to extract multi-scale point features of $\bm{P}$ and $\bm{Q}$, respectively. The output feature maps are denoted by $\bm{F}_p\in\mathbb{R}^{N_s\times C}$ and $\bm{F}_q\in\mathbb{R}^{M_s\times C}$, where $N_s$ and $M_s$ are the numbers of points after using grid subsampling~\cite{thomas2019kpconv}. After that, we simply use self-attention and cross-attention to build the correlation between the model point cloud and the scene point cloud, which is written as:

\begin{equation}
\begin{aligned}
\bm{F}_p = \operatorname{SelfAttn}(\bm{F}_p,&\bm{F}_p,\bm{F}_p),\bm{F}_q = \operatorname{SelfAttn}(\bm{F}_q,\bm{F}_q,\bm{F}_q),\\
\bm{H}_p &= \operatorname{CrossAttn}(\bm{F}_p,\bm{F}_q,\bm{F}_q)
\end{aligned}
\end{equation}

where $\bm{H}_p\in\mathbb{R}^{N\times C}$ is the output feature map of the scene point cloud $\bm{P}$ that embeds the relationship of the model point cloud $\bm{Q}$. Note that in the experiments, we stack three cross-attention layers, in which the output of the previous layer will be sent to the next layer as the input. For simplicity, we use the same symbol $\bm{H}_p$ to represent the feature map of the final output. By learning feature correlations, it is desired that the potential instances will be enhanced in the background, making them easier to detect.

\textbf{Object center prediction.} After feature correlation learning, we regress the object center from the whole scene. Given the feature map $\bm{H}_p\in\mathbb{R}^{N_s\times C}$, we directly predict the offset vector and the instance mask for each point. The point offset vector means the displacement to its instance center, which is given by:
\begin{equation}
{V}_p = \operatorname{MLP}(\bm{H}_p)
\end{equation}
where $\bm{V}_p\in\mathbb{R}^{N_s\times 3}$ is the offset matrix for $N_s$ points in the scene. To identify whether a point belongs to an instance rather than the background, we predict the point mask, which is written as:
\begin{equation}
\bm{Y}_p = \operatorname{MLP}(\operatorname{Concat}(\bm{H}_p, \bm{G}_p))
\end{equation}
where $\bm{Y}_p\in\mathbb{R}^{N_s\times 1}$ is the mask score. $\bm{G}_p$ is the geodesic distance embedding from~\cite{yu2024learning}. If the mask score is larger than 0.5, this point is identified as belonging to the point on the object. To obtain an accurate object center, we first displace each point to its potential instance center by adding the original coordinates of each point to its learned point offset ($\bm{P}+\bm{V}_p$). Then, we use the learned mask $\bm{Y}_p$ to filter out background points and leave points on the instance. Subsequently, we employ DBSCAN \cite{ester1996density} to group the offset points into K clusters. Finally, by averaging the points in each cluster, we can obtain the center of each instance, which is formulated by:
\begin{equation}
\bm{S}_p = \operatorname{Avg}(\bm{Y}_p\cdot\bm{V}_p)
\end{equation}
where $\bm{S}_p\in\mathbb{R}^{K\times 3}$ is the center of $K$ objects.

\textbf{Object proposal generation.} Based on the obtained object centers, we construct the object proposals through ball query operation. Based on the object center, we use radius $r$ to draw a three-dimensional sphere and collect the points that fall within the sphere as the object proposal. Note that the radius parameter $r$ is equal to the radius of the model point cloud. It is desired that the constructed spherical regions should include the entire object as much as possible. Compared with directly learning multi-instance point cloud registration, we can learn pair-wise point cloud registration between each object proposal and the model point cloud, thereby reducing the difficulty of registration.

\subsection{3D Dual-Masking Instance Matching}\label{sec:sec_3dmatch}
Once we obtain object proposals, we employ a 3D dual-masking instance matching module to learn pair-wise point cloud registration. Specifically, we first learn the instance mask to segment the instance from the object proposal. Then, we learn the overlap mask to segment the common area between the instance and the model point cloud. Finally, based on the instance mask and overlap mask, we learn the pair-wise instance matching.

\textbf{Instance mask.} Since we cannot obtain the ideal object proposal, we need to filter out the background points from the object proposal to obtain the mask of the instance. Given the point cloud of object proposal $\bm{O}\in\mathbb{R}^{T\times 3}$ ($T$ is the number of points in object proposal), we employ a small encoder structure of KPConv~\cite{thomas2019kpconv} to extract the feature of object proposal. Therefore, we can obtain the feature map $\bm{E}_o\in\mathbb{R}^{T_s\times C}$, where $T_s$ is the number of points after grid subsampling. The instance mask $\bm{Y}_o$ is formulated by:
\begin{equation}
\bm{Y}_o = \operatorname{MLP}(\operatorname{Concat}(\bm{E}_o, \bm{G}_o))
\end{equation}
where $\bm{Y}_o\in\mathbb{R}^{T_s\times 1}$ is the mask score and $\bm{G}_o$ is the geodesic distance embedding from~\cite{yu2024learning}. It is worth noting that in the first stage, we learn point masks for all instances from the entire scene, making it difficult to obtain accurate point masks for each instance. Here, we learn point masks from the object proposal, so we can obtain more accurate instance masks.

\textbf{Overlap mask.} Generally, due to object occlusion, there are a large number of incomplete objects. Therefore, we consider learning the overlap mask between the incomplete object and the complete model point cloud. We feed the model point cloud into the designed small KPConv encoder to obtain feature map $\bm{E}_q\in\mathbb{R}^{T_q\times C}$. Similarly, we use self-attention and cross-attention to learn the correlation between the object proposal and the model point cloud, which is formulated as:
\begin{equation}
\begin{aligned}
\bm{E}_o = \operatorname{SelfAttn}(\bm{E}_o,&\bm{E}_o,\bm{E}_o),\bm{E}_q = \operatorname{SelfAttn}(\bm{E}_q,\bm{E}_q,\bm{E}_q),\\
\bm{Z}_o &= \operatorname{CrossAttn}(\bm{E}_o,\bm{E}_q,\bm{E}_q)
\end{aligned}
\end{equation}
where $\bm{Z}_o\in\mathbb{R}^{T_s\times C}$ is the enhanced feature map of the object proposal. After that, we use an MLP to predict the overlap mask, which is given by:
\begin{equation}
\bm{Y}_{op} = \operatorname{MLP}(\operatorname{Concat}(\bm{Z}_o, \bm{G}_{o}))
\end{equation}
where $\bm{Y}_{op}\in\mathbb{R}^{T_s\times 1}$ is the obtained overlap mask. To upsample the overlap mask to the original resolution, we use a small KPConv decoder to generate feature map $\hat{\bm{Y}}_{op}\in\mathbb{R}^{T\times 1}$.

For subsequent matching steps, we follow~\cite{qin2022geometric} to match the dense points within the local patches of two matched sampled points with an optimal transport layer\cite{sarlin2020superglue}. However, the local correspondences extracted in this manner often cluster closely, which results in unstable pose estimation, as noted in\cite{predator}. Since the local area extracted from the focus network will contain some scene noise, and the intercepted instance point cloud will be incomplete, this problem will be more serious. To mitigate this problem, we suggest extracting a dense set of point correspondences within the instance boundaries by utilizing instance masks and overlap masks. For each points correspondences $\hat{\mathcal{C}}_{k}=(\hat{\mathbf{p}}_{i},\hat{\mathbf{q}}_{j})$, we collect their neighboring points $\mathcal{N}_{i}^{P}$ and $\mathcal{N}_{j}^{Q}$. The points out of instance is removed from $\mathcal{N}_{i}^{P}$ based on the instance mask. In order to solve the problem of incomplete point clouds, we further use overlap masks to eliminate non-overlapping parts within a patch. Then the clear pair-wise correspondences are extracted with an optimal transport layer and mutual top-k selection, and using the local-to-global registration followed by \cite{qin2022geometric}.

\subsection{Loss Function}\label{sec:sec_loss}
\textbf{Loss in focusing.} For the 3D multi-object focusing module, we need to learn better shape features for localization. Therefore, we follow \cite{qin2022geometric} and use a circle loss $L_{circle}$ to learn fine interactive features, as:
\begin{equation}
L_{\mathrm{circle}}^{Q}=\frac{1}{|\mathcal{A}|}\sum_{\mathcal{G}_{i}^{Q}\in\mathcal{A}}\log[1+\sum_{\mathcal{G}_{j}^{P}\in\varepsilon_{p}^{i}}e^{\lambda_{i}^{j}\beta_{p}^{i,j}(d_{i}^{j}-\Delta_{p})}\cdot\sum_{\mathcal{G}_{k}^{P}\in\varepsilon_{n}^{i}}e^{\beta_{n}^{i,k}(\Delta_{n}-d_{i}^{k})}]
\end{equation}
where P and Q are source and target points-set and  $\mathcal{G}$ is the anchor patches of each set. $d_{i}^{j}=\|\hat{\mathbf{h}}_{i}^{Q} - \hat{\mathbf{h}}_{j}^{P}\|_{2}$ is the distance in feature space, $\lambda_i^j=(o_i^j)^{\frac{1}{2}}$ and $o_i^j$ is the overlap ratio between $\mathcal{G}_i^{P}$ and $\mathcal{G}_j^{Q}.$ The weights $\beta_{p}^{i,j}=\gamma(d_{i}^{j}-\Delta_{p})$ and $ \beta_{n}^{i,k}=\gamma(\Delta_{n}-d_{i}^{k})$ are determined individually for each positive and negative example, using the margin hyper-parameters $\Delta_p=0.1$ and $\Delta_n=1.4$. The circle loss on $\mathcal{P}$ is calculated in the same way.\\
For each sampled points in the scene, we constrain their learned offsets $\textbf{O}=\{o_1,...,o_N\}\in \mathbb{R}^{N\times3}$ from their nearest target instance center using an L1 regression loss as follows:
\begin{equation}
L_{reg} = \frac{1}{\sum_{i}p_i}\sum_i \| \mathbf{\mathit{o}_\mathit{i}-(\hat{\mathit{c}_\mathit{i}}-\mathit{p}_\mathit{i})} \|
\end{equation}
where $\hat{c_i}$ is the centroid of the nearest instance that points $\mathit{i}$ belongs to. Considering the varying object sizes across different categories, it is challenging for the network to accurately regress precise offsets, especially for boundary points of large objects, as these points are relatively far from the instance centroids. To tackle this problem, we introduce a direction loss to constrain the direction of the predicted offset vectors. We define this loss, following the method in\cite{jiang2020pointgroup}, as a measure of the negative cosine similarities, $\mathit{i.e.}$,
\begin{equation}
L_{dir} = -\frac{1}{\sum_{i}p_i}\sum_i \frac{\mathit{o}_\mathit{i}}{\| \mathbf{\mathit{o}_\mathit{i}}\|_2}\cdot\frac{\hat{\mathit{c}_\mathit{i}}-\mathit{p}_\mathit{i}}{\| \mathbf{\hat{\mathit{c}_\mathit{i}}-\mathit{p}_\mathit{i}} \|_2} 
\end{equation}
The overall focusing loss is computed as: $L_{focusing}=L_{circle}+L_{reg}+L_{dir}$.

\textbf{Loss in matching.} For coarse points feature learning, we employ the circle loss, as mentioned earlier in the Focus Network. As for point matching, we following \cite{qin2022geometric} use a negative log-likelihood loss on the assignment matrix $\mathbf{\bar{Z_\mathit{i}}}$ of each ground-truth points correspondence $\hat{c}_i^*$, just as following:
\begin{equation}
L_{\mathrm{nll},i}=-\sum_{(x,y)\in\mathcal{C}_{i}^{*}}\log\bar{z}_{x,y}^{i}-\sum_{x\in\mathcal{I}_{i}}\log\bar{z}_{x,m_{i}+1}^{i}-\sum_{y\in\mathcal{J}_{i}}\log\bar{z}_{n_{i}+1,y}^{i}
\end{equation}
where $\mathcal{I}_{i}$and$\mathcal{J}_{i}$ are the unmatched points in the two
matched patches. The final loss is the average of the loss over all points matches: $L_{\mathrm{nll}}=\frac1{N_g}\sum_{i=1}^{N_g}L_{p,i}.$\\
Regarding the prediction of instance masks and overlap masks, we follow the methodology outlined in \cite{milletari2016v}. The mask prediction loss is composed of binary cross-entropy (BCE) loss and dice loss with Laplace smoothing, defined as follows:
\begin{equation}
L_{\text{mask},i}=\text{BCE}(m_i,m_i^{gt})+1-2\frac{m_i\cdot m_i^{gt}+1}{|m_i|+|m_i^{gt}|+1}
\end{equation}
where $m_i$ and $m_i^{gt}$are the predicted and the ground-truth instance masks, respectively. The final mask prediction loss is the average loss over all.
The total matching loss function $L_{matching}$ is defined as:
\begin{equation}
L_{matching}=L_{circle}+L_{nll}+L_{overlapmask}+L_{instancemask}.  
\end{equation}

\section{Experiments}
\subsection{Datasets and Evaluation Metrics}
For multi-instance point cloud registration, We train and evaluate our method on two public benchmarks: Scan2CAD~\cite{avetisyan2019scan2cad} and ROBI~\cite{yang2021robi}.

\textbf{Scan2CAD}. As a pioneering dataset in the realm of aligning scenes with CAD models, it leverages the resources of ScanNet~\cite{scannet} and ShapeNet~\cite{chang2015shapenet} to form a multi-instance registration dataset. With a corpus comprising 1,506 scenes sourced from ScanNet, meticulously annotated with 14,225 CAD models from ShapeNet alongside their spatial orientations within the scenes, Scan2CAD sets a new standard in scene-CAD alignment. For the total of 2,184 pairs of point clouds, it allocates 70\% of pairs for training, 10\% for validation, and reserves 20\% for testing.

\textbf{ROBI}. It is a dataset tailored specifically for industrial bin-picking applications. ROBI collects 7 reflective metallic industrial objects and 63 meticulously crafted bin-picking scenes.  Each point cloud pair is meticulously crafted, with the scene point cloud generated through back projection from depth images, while the model point cloud is meticulously sampled from the CAD model corresponding to its industrial counterpart. There are a total of 4,880 pairs of ROBI, divided into 70\% for training, 10\% for validation, and 20\% for testing.

\textbf{Metrics}. We adopt three registration metrics to evaluate the methods, including Mean Recall (MR), Mean Precision (MP), and Mean F$_1$ score (MF). We refer to the settings used in MIRETR and previous work to determine whether an instance is recognized as correctly registered based on RTE and RRE \cite{yuan2022pointclm,tang2022multi}. Specifically, we consider a match successful when $RTE \leq 4 \times voxelsize$ and $RRE \leq 15^\circ$. Following existing methods, such as MIRETR\cite{yu2024learning}, the voxel sizes of Scan2CAD and ROBI dataset are set to 0.025m and 0.0015m, respectively. MR quantifies the proportion of registered instances relative to the total number of ground-truth instances, while MP measures the ratio of registered instances against the entirety of predicted instances. MF score is the harmonic mean of both MP and MR. Additionally, we present the pair-wise inlier ratio (PIR), elucidating the proportion of inliers from a single instance amidst all extracted correspondences, considering one of the most important challenges of multi-instance registration is identifying the set of inliers for individual instances.

\subsection{Implementation Details}
Our method is trained on NVIDIA RTX 4090 GPUs and uses the Pytorch deep learning platform. We employ the Adam optimizer for 60 epochs. In initial learning rate and weight decay are set to 0.001 and 0.0001, respectively. We use a KPConv-FPN\cite{thomas2019kpconv} backbone followed by \cite{yu2024learning} for feature extraction. We utilize a voxel subsampling approach to reduce the resolution of the point clouds, resulting in the creation of sampled points and dense points, which are then inputted into the network. The initial step involves downsampling the input point clouds using a voxel-grid filter with a size of 2.5cm for Scan2CAD and 0.15cm for ROBI. Subsequently, we employ a 4-stage backbone architecture in both the multi-object focusing and sub-matching network. Following each stage, the voxel size is increased twofold to further reduce the resolution of the point clouds. The initial and final (coarsest) levels of downsampled points correspond to the dense points and sampled points, respectively, which are used for follow-up process. In the multi-object focusing network, we employ a ball query approach akin to the one detailed in \cite{pointnet++}, enabling us to retrieve the local neighborhood surrounding the regression center. The search radius is set to 1.2 times the size of the CAD model. Specifically, we randomly sample 4096 points from the dense points obtained through voxel subsampling in both Scan2CAD and ROBI datasets. During training, we use the ground truth center as supervision to train the 3D multi-object focusing module and use the point cloud around the ground truth center as the training data to train the matching network. During testing, we use the center predicted by the 3D multi-object focusing module and its surrounding point cloud as the input to the 3D dual-masking instance matching module to regress the final pose.

\begin{table}[t]
    \caption{Results comparison of different methods on the test sets of both the Scan2CAD and ROBI datasets. The best results are highlighted in \textbf{bold}. Please note that ``3DFMNet$^*$'' indicates the upper bound of our method.}
\label{tab:results_scan2cad_robi}
\centering
\resizebox{0.8\textwidth}{!}{
    \begin{tabular}{c|ccc|ccc}
        \toprule
        \multirow{3}{*}{Methods} &
        \multicolumn{3}{c|}{Scan2CAD} & \multicolumn{3}{c}{ROBI}\\ 
        \cmidrule(r){2-7} & MR (\%) & MP (\%) & MF (\%) & MR (\%) & MP (\%) & MF (\%)\\
        \midrule
        T-Linkage~\cite{magri2014t} & 77.12 & 46.04 & 57.65 & 12.04 & 10.47 & 11.20\\
        RansaCov~\cite{magri2016multiple} & 84.78 & 71.34 & 77.48 & 14.14 & 26.29 & 18.38\\
        PointCLM~\cite{yuan2022pointclm}  & 91.85 & 91.08 & 91.46 & 18.68 & 40.11 & 25.48\\
        ECC~\cite{tang2022multi} & \textbf{96.52} & 89.03  & 92.62 & 24.65 & 34.85 & 28.91\\
        MIRETR~\cite{yu2024learning} &  95.70 & 91.21 & 93.40 & 38.51 & 41.19 & 39.80\\
        \midrule
        3DFMNet (ours) & 95.44 & \textbf{94.15} & \textbf{94.79} & \textbf{46.81} & \textbf{50.61} & \textbf{48.63} \\
        \textcolor{gray}{3DFMNet$^*$ (ours)} & \textcolor{gray}{97.68} & \textcolor{gray}{94.63} & \textcolor{gray}{96.14}
        & \textcolor{gray}{52.59} & \textcolor{gray}{63.13} & \textcolor{gray}{57.38} \\
        \bottomrule
\end{tabular}}
\end{table}

\subsection{Results}\label{sec:sec_results}

\textbf{Quantitative results.} For a comprehensive comparison, we compare our 3DFMNet with five recent advanced multi-instance point cloud registration approaches, including T-Linkage~\cite{magri2014t}, RansaConv~\cite{magri2016multiple}, PointCLM~\cite{yuan2022pointclm}, ECC~\cite{tang2022multi}, and MIRETR~\cite{yu2024learning}. As shown in Table~\ref{tab:results_scan2cad_robi}, we report the results of different methods on Scan2CAD~\cite{avetisyan2019scan2cad} and ROBI~\cite{yang2021robi}, respectively. Note that except for the results of our method, the results of all other methods are taken from the official paper of MIRETR, as their results are obtained using the same backbone network. From the table, it can be observed that the performance of our method is superior to all other methods on the ROBI datasets. The density of objects in ROBI is higher than that in Scan2CAD, making ROBI more challenging. It can be observed that our method improves performance by about 7\% in terms of MR, MP, and MF on the ROBI dataset. Since we use the focus-and-matching strategy to learn multiple pair-wise registrations, our method can learn more accurate matches from object proposals compared to the whole scene.

Since our method first localizes the object and then executes pair-wise correspondence, the precision of object localization has a significant impact on subsequent matching. In Table~\ref{tab:results_scan2cad_robi}, we provide the upper bound on the performance of our method, denoted by ``3DFMNet$^*$''. Specifically, we utilize the ground truth object center to generate the object region, and use it to evaluate the performance upper bound of our method. For the Scan2CAD dataset, the upper bounds of MR and MP are about 97\% and 94\%, respectively. However, due to the cluttered and incomplete objects in the ROBI dataset, the upper bounds for MR and MP are only about 52\% and 63\%. Nonetheless, the performance upper bound of our method is still higher than previous methods. In a word, the theoretical upper bound indicates that the strategy of focusing first and then matching can ensure that our method can achieve excellent performance.

To evaluate the 3D multi-object focusing module, we computed the mean recall (MR), mean precision (MP), and root mean square error (RMSE) for detected object centers. We defined successful detection as cases where the predicted center lies within $distance \leq 0.1\times r_{instance}$ (where $r_{instance}$is the radius of instance) of the ground truth center. As shown in Table~\ref{tab:3D_focusing_module}, the module achieves strong results across MR, MP, and RMSE, confirming its effectiveness in identifying object centers in the first stage. n the second stage, the 3D dual-masking instance matching module refines detections by applying instance and overlap mask learning to reduce false positives. Analyzing the Scan2CAD dataset, we found that 1.05\% of objects were initially misdetected (25 instances), but 22 were successfully filtered out by the masking module. These misdetections have minimal impact, as few matching points are generated for falsely detected objects, preventing SVD from calculating a relative pose and limiting their influence on final results.

For unseen scene experiments, we follow MIRETR by using the ShapeNet dataset \cite{chang2015shapenet} (55 categories in total) to evaluate generalization to novel categories. Specifically, we train on CAD models from the first 30 categories and test on the remaining 25. To address class imbalance, we sample up to 500 models per category, as in MIRETR. Each point cloud pair includes a randomly selected CAD model and a scene model with 4–16 random poses applied, resulting in 8,634 pairs for training, 900 for validation, and 7,683 for testing. Table~\ref{tab:unseen_scenes} shows that our method demonstrates strong generalizability on unseen scenes compared to MIRETR.

\begin{figure}[h]
    \centering
    \begin{minipage}{0.48\textwidth}
        \centering
        \captionof{table}{Result related to the 3D multi-object focusing module.}
        \label{tab:3D_focusing_module}
        \resizebox{\textwidth}{!}{%
        \begin{tabular}{cccc}
            \toprule
            Datasets & MR (\%) & MP (\%) & RMSE (m)  \\
            \midrule
            Scan2CAD~\cite{avetisyan2019scan2cad} & 98.14 & 98.85 & 0.0814 \\
            ROBI~\cite{yang2021robi} & 80.30 & 99.99 & 0.0065 \\
            \bottomrule
        \end{tabular}}
    \end{minipage}%
    \hfill
    \begin{minipage}{0.48\textwidth}
        \centering
        \captionof{table}{The generalizability of the 3D multi-object focusing module to unseen scenes.}
        \label{tab:unseen_scenes}
        \resizebox{\textwidth}{!}{%
        \begin{tabular}{cccc}
            \toprule
            Methods & MR (\%) & MP (\%) & MF (\%)  \\
            \midrule
            MIRETR~\cite{yu2024learning} & 94.95 & 93.94 & 94.44 \\
            3DFMNet (ours) & \textbf{95.12} & \textbf{94.23} & \textbf{94.67} \\
            \bottomrule
        \end{tabular}}
    \end{minipage}
\end{figure}

\begin{figure}[h]
    \centering
    \includegraphics[width=0.95\textwidth]{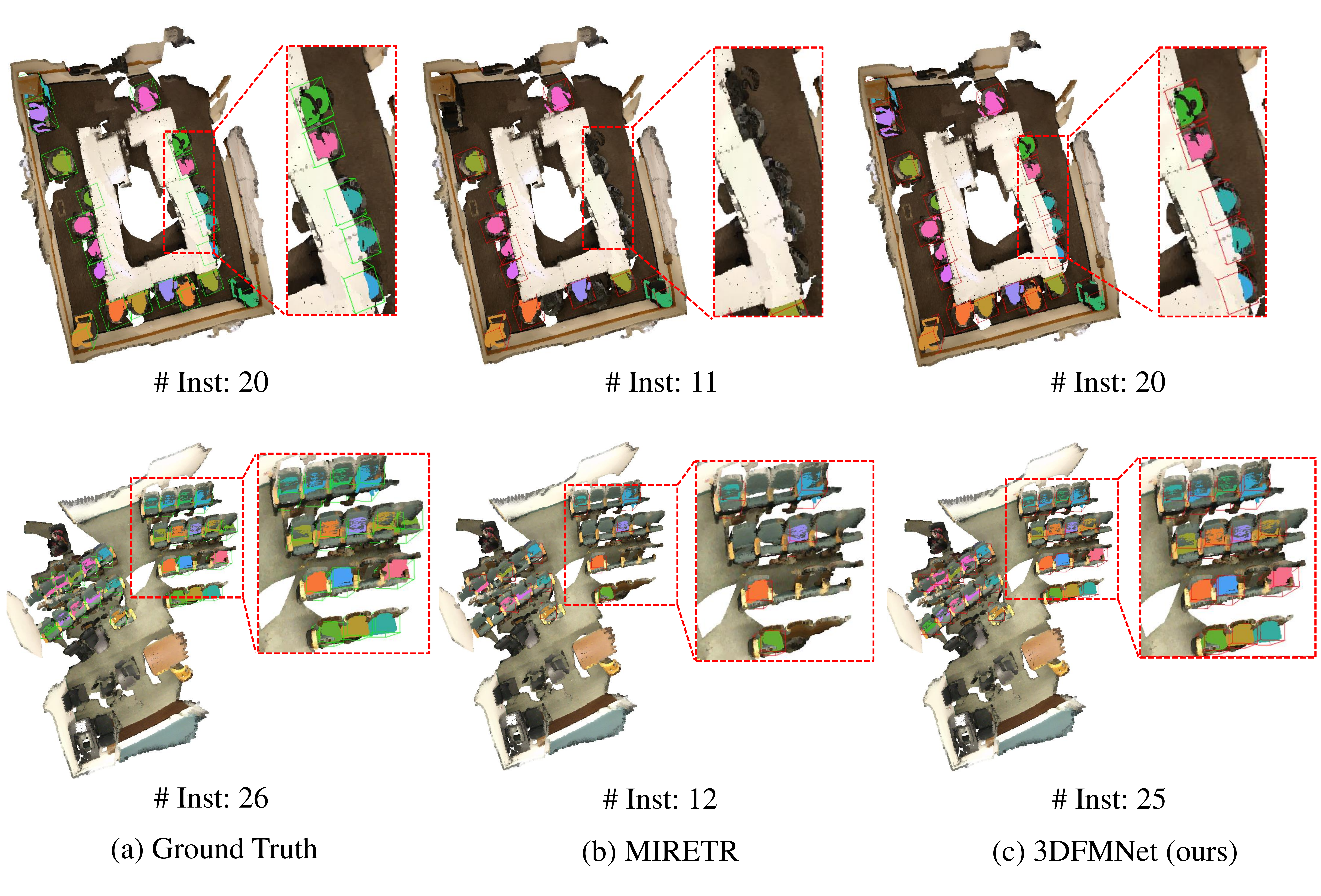}
    \caption{Registration results on the test set of the Sacn2CAD dataset. We visualize the successfully registered instances of MIRETR~\cite{yu2024learning} in (b) and ours in (c). ``\# Inst'' means the number of registered instances. Note that for a better view, we draw the \textcolor{green}{green boxes} for the ground truth and the \textcolor{red}{red boxes} for the predict correspondences.}
    \label{fig:vis_scan2cad}
\end{figure}

\begin{figure}[h]
    \centering
    \includegraphics[width=\textwidth]{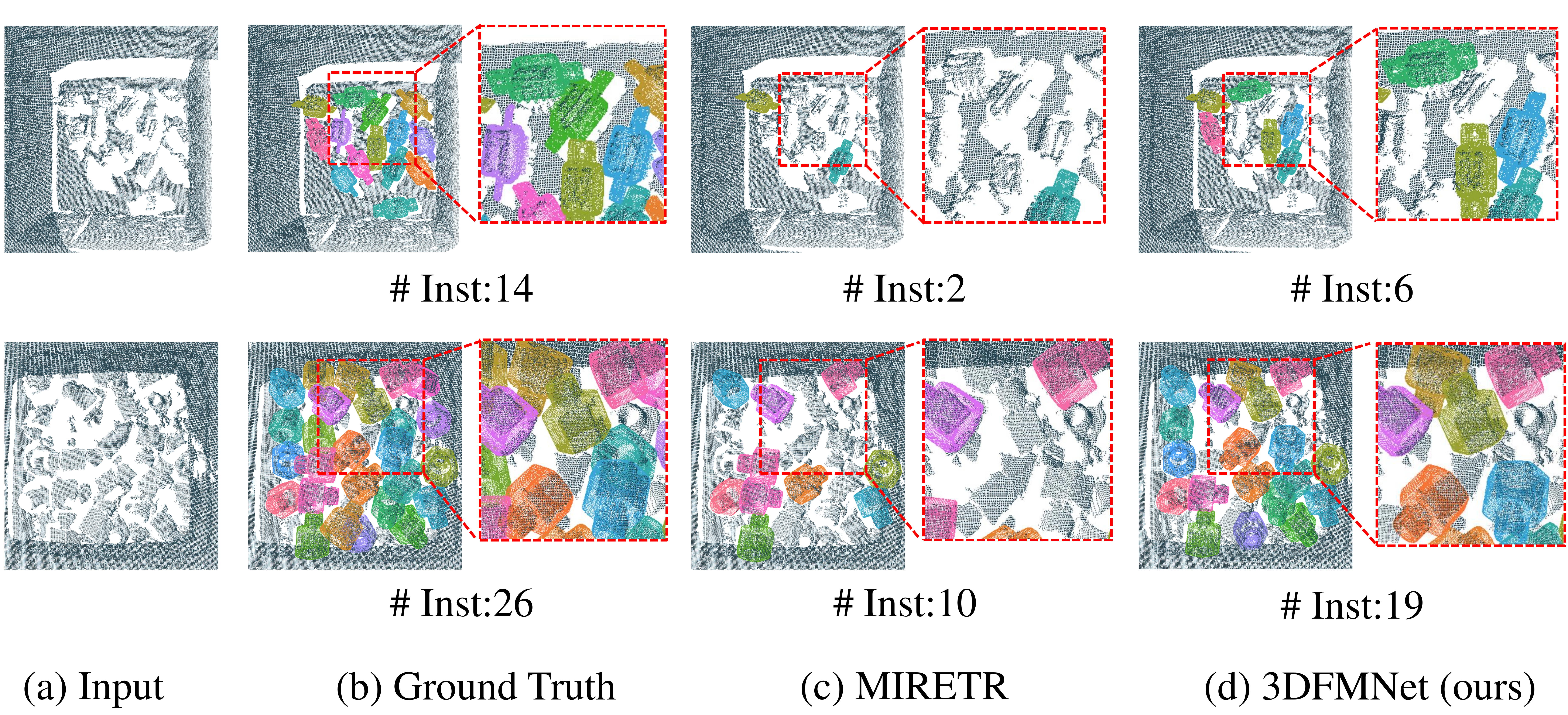}
    \caption{Registration results on the test set of the ROBI dataset. We visualize the successfully registered instances of MIRETR~\cite{yu2024learning} in (c) and ours in (d).}
    \label{fig:vis_robi}
\end{figure}

\begin{figure}[h]
    \centering
    \includegraphics[width=\textwidth]{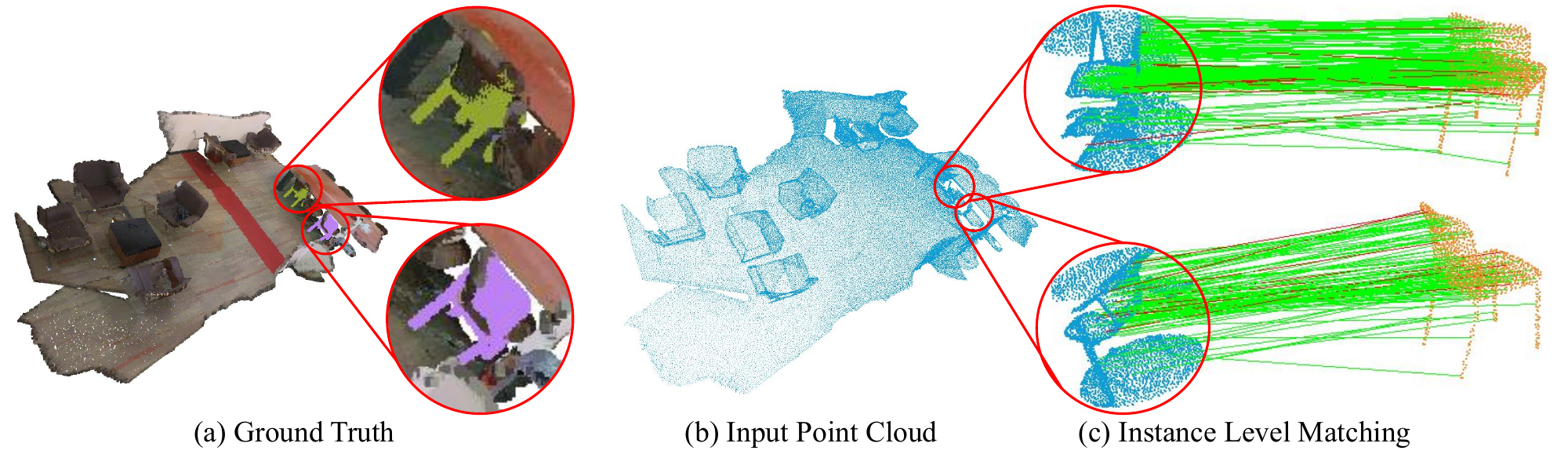}
    \caption{Visualization results of pair-wise correspondences on the test set of Scan2CAD dataset.}
    \label{fig:vis_inlier}
\end{figure}

\textbf{Visualization.} In Figure~\ref{fig:vis_scan2cad}, we show the visualization of chairs of our method with the previous state-of-the-art MIRETR~\cite{yu2024learning} on the test set of the Scan2CAD dataset. From the left to right, the figure shows the results of ground truth, MIRETR, and 3DFMNet (ours), respectively. Note that we transform the chair's CAD model to the corresponding scenes. By comparing the ground truth with our method, it can be observed that our method can successfully match the multiple chairs in the cluttered scene. In addition, we also show the pair-wise correspondences on the test set of the Scan2CAD dataset in Figure~\ref{fig:vis_inlier}. In Figure~\ref{fig:vis_robi}, we also show the visualization of the incomplete objects on the test set of the ROBI dataset. It can be observed that our method can effectively obtain the correspondences between the challenging incomplete objects and the model point cloud.

\begin{wrapfigure}{r}{0.45\textwidth}
     \vspace{-12pt}
     \captionof{table}{Per scene time on the ROBI dataset.}
     \label{tab:time_cmp}
     \centering
     \resizebox{0.45\textwidth}{!}{%
     \begin{tabular}{cccc}
         \toprule
         Methods & Model & Pose & Total  \\
         \midrule
         T-Linkage~\cite{magri2014t} & 0.30 &  3.07 & 3.34 \\
         RansaCov~\cite{magri2016multiple} & 0.30 & 0.17 &  0.47 \\
         PointCLM~\cite{yuan2022pointclm} & 0.30 & 0.33 &  0.63\\
         ECC~\cite{tang2022multi} & 0.30 &  0.21 & 0.51 \\
         MIRETR~\cite{yu2024learning} & 0.30 &  \textbf{0.10} & \textbf{0.40} \\
         3DFMNet (ours) & \textbf{0.26} & 0.28 & 0.54\\
         \bottomrule
     \end{tabular}}
\end{wrapfigure}

\textbf{Time costs.} Here we report the inference time of different methods on the ROBI dataset for a comprehensive comparison. Table~\ref{tab:time_cmp} shows model inference time for feature extraction and pose estimation time for transformation. ``Total'' represents the sum of ``Model'' time and ``Pose'' time. While our two-stage method has a slightly higher total time than the one-stage MIRETR~\cite{yu2024learning}, it runs faster than the two-stage PointCLM~\cite{yuan2022pointclm}. Nonetheless, compared with PointCLM and MIRETR, our 3DFMNet achieves higher performance.


\subsection{Ablation Study}

\textbf{Impact of the number of sampled points.} The grid sizes influence the number of sampled points. We conduct experiments on the Scan2CAD dataset to verify the impact of different numbers of points. The results of MR and MP are 83.76\% and 92.50\% (1024 points), \textbf{95.44}\% and \textbf{94.15}\% (default 4096 points), and 95.20\% and 93.75\% (8192 points), respectively. It is evident that too few sampling points hinder information gathering and reduce accuracy. Conversely, too many sampling points do not improve accuracy and instead decrease it due to redundancy.

\textbf{Effective of dual-masking.} We evaluate the necessity of our dual-masking structure under the fair setting of the Scan2CAD dataset. When taking turns removing components, the results of MR and MP are 94.76\% and 93.30\% (only removing the overlap mask), 91.82\% and 93.53\% (only removing the instance mask), 90.01\% and 90.90\% (removing both), and \textbf{95.44}\% and \textbf{94.15}\% (using both). The ablation study results can demonstrate the effectiveness of our dual-masking structure.

\subsection{Limitation}
Our 3D focusing-and-matching network is a two-stage framework for the multi-instance point cloud registration task. The localization accuracy of the first stage will affect the pair-wise correspondence in the second stage. We have analyzed the upper bound of our method in \ref{sec:sec_results}. In addition, due to the two-stage process, the inference time of our method is slightly lower than the previous MIRETR~\cite{yu2024learning}. In future work, we will strive to improve the performance and speed of our method.

\section{Conclusion}
In this paper, we proposed a 3D focusing-and-matching network (3DFMNet) for multi-instance point cloud registration. Specifically, we first presented a 3D multi-object focusing module that learns to localize the center of the potential target in the scene by considering the correlation between the model point cloud and the scene point cloud. Then, we designed a 3D dual-masking instance matching module to learn the pair-wise correspondence between the model point cloud and the localized object. Extensive experiments on two public benchmarks, Scan2CAD and ROBI, show that our method achieves a new state-of-the-art performance on the multi-instance point cloud registration task.

\section*{Acknowledgments} The authors would like to thank reviewers for their detailed comments and instructive suggestions. This work was supported by the National Science Fund of China (Grant Nos. 62306238, 62271410, 62001394) and the Fundamental Research Funds for the Central Universities.

\bibliographystyle{plain}
\small\bibliography{neurips_2024}

\end{document}